\documentclass[10pt, a4paper, conference, compsocconf, twocolumn]{IEEEtran}
\usepackage{ textcomp }

\usepackage{hyperref}       

\usepackage{url}            

\usepackage{booktabs}       

\usepackage{amsfonts}       

\usepackage{nicefrac}       

\usepackage{microtype}      

\usepackage{graphicx}

\usepackage{subcaption}

\usepackage{amsmath}

\usepackage{algpseudocode}

\usepackage{algorithm}
	
\algnewcommand\algorithmicforeach{\textbf{for each}}

\algdef{S}[FOR]{ForEach}[1]{\algorithmicforeach\ #1\ \algorithmicdo}

\begin{document}
	
	\title{Fine-tuning Handwriting Recognition systems with Temporal Dropout}
	
	\author{
		\IEEEauthorblockN{Edgard Chammas}
		\IEEEauthorblockA{RNNLab\\Beirut, Lebanon\\edgard@rnnlab.com}
		\and
		\IEEEauthorblockN{Chafic Mokbel}
		\IEEEauthorblockA{University of Balamand\\El-Koura, Lebanon\\chafic.mokbel@balamand.edu.lb}
	}

	\maketitle
	
	\begin{abstract}
		This paper introduces a novel method to fine-tune handwriting recognition systems based on Recurrent Neural Networks (RNN). Long Short-Term Memory (LSTM) networks are good at modeling long sequences but they tend to overfit over time. To improve the system's ability to model sequences, we propose to drop information at random positions in the sequence. We call our approach Temporal Dropout (TD). We apply TD at the image level as well to internal network representation. We show that TD improves the results on two different datasets. Our method outperforms previous state-of-the-art on Rodrigo dataset.
	\end{abstract}
	
	
	\IEEEpeerreviewmaketitle
	
	\section{Introduction}
	\label{section1}
	Dropout is a regularization technique that is often used to prevent neuron co-adaptation which reduces overfitting and improves network generalization. The technique consists of randomly disabling individual neurons by masking out their activations (output weights) \cite{hinton2012improving}. A generalization of dropout, DropConnect, works by randomly disabling individual neuron weights \cite{wan2013regularization}. Dropout was primarily intended to be applied to intermediate layers of the network. However, it could also be applied in the input space. For instance, DeVries et al. \cite{devries2017improved} improves the accuracy of a convolutional neural network (CNN) for image classification by randomly masking out a region of the input image. This technique can be regarded as an extension of dropout, but with a spatial prior applied. The main motivation behind such a technique is to improve the model's ability to handle object occlusion and take context into consideration. Even though not directly applied to the input layer, the term Spatial Dropout first appeared in the work of Tompson et al. \cite{tompson2015efficient}, which proposed to improve the performance of CNN by randomly discarding entire feature maps instead of pixels.
	
	In this work, we propose a new regularization technique in the form of a dropout layer, to improve the performance of a convolution recurrent neural network (CRNN) for unconstrained offline handwriting recognition. Our approach exploits the sequential mechanism that is used by most Handwritten Text Recognition (HTR) systems to process text-line images. More generally, we hypothesize that any system working on sequential data could be improved by stochastically removing (dropping out) some elements from the sequence during training. This would help the system to better model contextual information and improve its robustness toward intra-sequence variability. We propose to fine-tune our recognition system by adding a dropout layer; we call it Temporal Dropout (TD). We describe this approach in details in section \ref{section2} and show how it can be used at different layers of the HTR system. In section \ref{section3}, we assess network co-adaptation with a variation of the TD approach that does not involve information loss. We present our baseline HTR system and the two datasets that we used in sections \ref{section4} and \ref{section5} respectively. Finally, we discuss the different experiments conducted in section \ref{section6} where we assess the effect of TD on the HTR system.

\section{Temporal Dropout}
\label{section2}

\begin{figure}[h]
	\centering
	\includegraphics[width=1.0\linewidth]{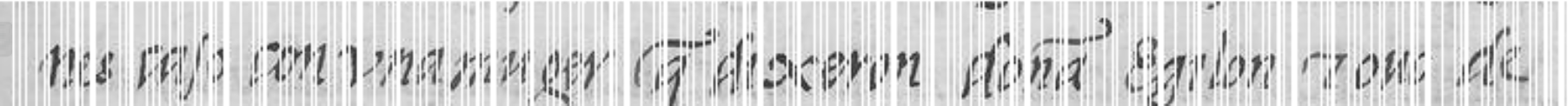}
	\caption{Input text-line image with 30\% temporal dropout.}
	\label{figure1}
\end{figure}

Most handwriting recognition systems (working at the line level) use a sliding window approach for feature extraction. The window is swiped along the horizontal axis, usually in the direction of writing (e.g., left to right for Latin handwriting), to extract specific characteristics of the script. At each position of the window, features are extracted as a vector, and then concatenated to a sequence. This spatio-temporal mapping from image pixels to feature sequence is achieved through an encoder network (usually a CNN). The features sequence is then modeled by a decoder network (usually an RNN). In the input space, our approach is similar to that of DeVries et al. \cite{devries2017improved} which randomly removes a square-shaped region from the input image. Instead, we propose to remove complete column regions at random positions of the input image (see figure \ref{figure1}). Considering the spatio-temporal mapping performed during feature extraction, hence we named our approach Temporal Dropout (TD). That should not be confused with the temporal dropout technique applied to video frames for spatio-temporal feature learning \cite{culibrk2014temporal}.

\begin{figure}[h]
	\centering
	\includegraphics[width=0.6\linewidth]{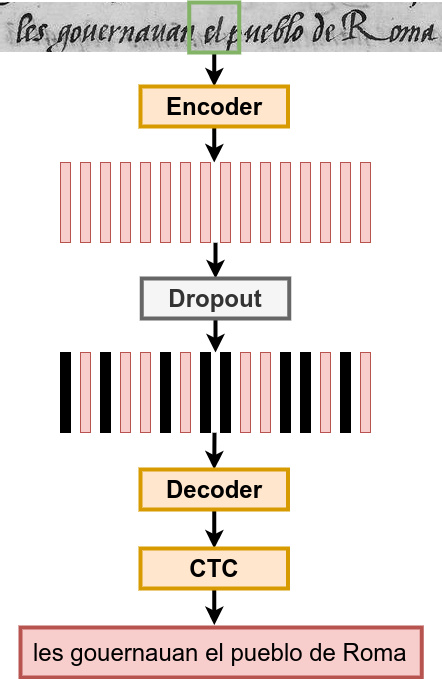}
	\caption{HTR recognition pipeline with temporal dropout applied to the CNN encoder output. The example illustrate a 50\% dropout rate where the dropped feature vectors are represented in black color.}
	\label{figure2}
\end{figure}

When applied to the input space, our approach could be seen more as a data augmentation method rather than a dropout. However, we do extend this approach to intermediate layers of the network. More specifically, we perform TD on the CNN encoder output (input of the RNN decoder) by removing (zeroing) vectors at random positions in the feature sequence (see figure \ref{figure2}). More formally, a Long Short-Term Memory (LSTM) network can be represented by the following equations:

\begin{equation}
	\begin{pmatrix}
	i_t \\
	f_t \\
	o_t \\
	g_t \\
	\end{pmatrix} = \begin{pmatrix}
	\sigma(W_i[x_t, h_{t-1}] + b_i) \\
	\sigma(W_f[x_t, h_{t-1}] + b_f) \\
	\sigma(W_o[x_t, h_{t-1}] + b_o) \\
	\sigma(W_g[x_t, h_{t-1}] + b_g) \\
	\end{pmatrix}
\end{equation}

\begin{equation}
	c_t = f_t * c_{t-1} + i_t * g_t
\end{equation}

\begin{equation}
	h_t = o_t * f(c_t)
\end{equation}

where $i_t$, $f_t$ and $o_t$ are the input, forget and output gates at time step $t$ respectively; $g_t$ is the cell update vector while $c_t$ and $h_t$ are the updated cell vector and the hidden state respectively. We denote by $\sigma$ the sigmoid activation function and $*$ the element-wise multiplication operator. Introducing a dropout mechanism to recurrent neural networks could be addressed in many ways. A direct application of dropout to handwriting recognition is the work of \cite{pham2014dropout} where the authors chose to apply dropout to feed forward connections only, restricting its use to input-hidden and hidden-output connections. The authors claim that one shall not apply dropout to recurrent connections as it would hurt the RNN's ability to model sequences. On the other hand, \cite{gal2016theoretically} proposed to apply dropout on the previous hidden state $h_{t-1}$ while \cite{moon2015rnndrop} and \cite{semeniuta2016recurrent} chose to apply it either directly on the cell values $c_{t-1}$, or on the cell update vector $g_t$ respectively. In this work, we propose to apply dropout directly on the input sequence. At each time step, equation 1 could be rewritten as:

\begin{equation}
\begin{pmatrix}
i_t \\
f_t \\
o_t \\
g_t \\
\end{pmatrix} = \begin{pmatrix}
\sigma(W_i[d(x_t), h_{t-1}] + b_i) \\
\sigma(W_f[d(x_t), h_{t-1}] + b_f) \\
\sigma(W_o[d(x_t), h_{t-1}] + b_o) \\
\sigma(W_g[d(x_t), h_{t-1}] + b_g) \\
\end{pmatrix}
\end{equation}

where $d$ is the dropout function defined as:

\begin{equation}
d(x_t)= 
\begin{cases}
m * x_t,& \text{if } train\ phase\\
x_t,              & \text{otherwise}
\end{cases}
\end{equation}

In contrast to the previous approaches where the activation vector is partially dropped, we chose to drop all the activations at once. The input vector is multiplied by a constant $m$ sampled, for each time step, from a Bernoulli distribution with success probability $p$. Because all vector values are either kept or dropped, no scaling is required. Compared to the other methods our approach can be seen as a special case with the same analogy as the relation between dropout (where entire columns of the weights matrix are dropped) and drop-connect. 

\section{Complementary Input Representations}
\label{section3}
In general, the Temporal Dropout approach falls under Noise Injection Regularization Techniques (NIRT), which have shown to improve the performance of neural networks \cite{an1996effects}\cite{neelakantan2015adding}. Network regularization could take different forms. For instance, one could target network weights by injecting adaptive Gaussian noise \cite{li2016whiteout} or by applying random changes to the weights \cite{kang2017shakeout}. At a higher level, some methods aim to regularize the network architecture. For instance, \cite{larsson2016fractalnet} regularizes sub-network paths by randomly dropping operands of the join layers. Whereas \cite{huang2016deep} performs network depth regularization as a way to solve the problem of vanishing gradients for very deep networks.

\begin{figure}[h]
	\centering
	\includegraphics[width=0.85\linewidth]{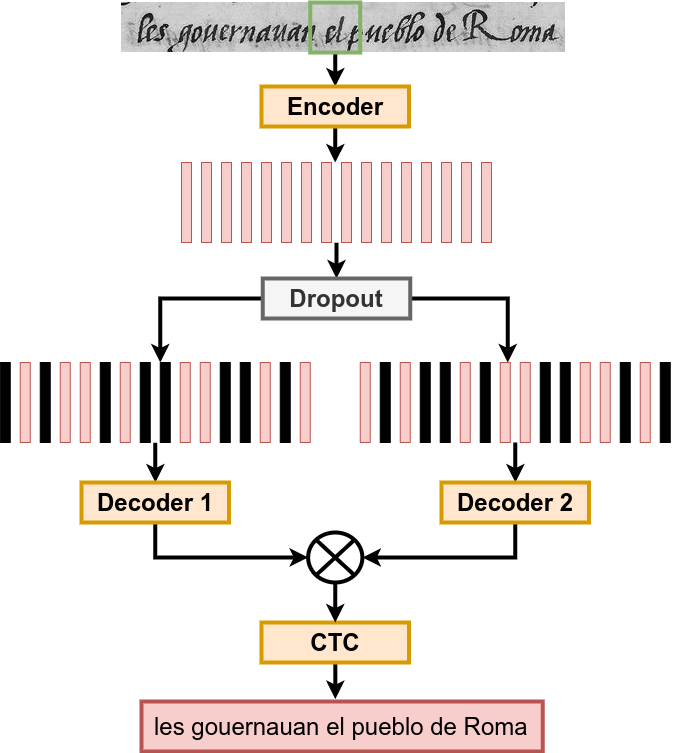}
	\caption{HTR recognition pipeline with 50\% temporal dropout applied to the CNN encoder output. Dropped feature vectors (represented in black) are fed as complementary input to another RNN decoder. We add the output of both decoders to produce the final transcription.}
	\label{figure3}
\end{figure}

Inspired by the work of \cite{gastaldi2017shake}, which attempts to apply data augmentation to internal representations, we propose to add a complementary RNN decoder ($RNN_{comp}$) to which we feed the dropped out feature vectors of the CNN encoder (see figure \ref{figure3}). We end up having two RNNs working in parallel, processing complementary input representations. The final output of the system is a sum of both networks:
\begin{equation}
y_t = RNN(m * x_t) + RNN_{comp}((1 - m) * x_t)
\end{equation}

In contrast to the TD approach presented in the previous section, this time, there is no loss of information. Despite having partial feature sequences presented to each RNN decoder, the total information is preserved within the network. Considering that each RNN decoder sees a complementary version of the input, hence we name this approach Temporal Dropout with Complementary Input Representation. The main intuition behind this approach is to assess the performance of TD without loss of information, as well as to assess the RNN decoders co-adaptation with respect to this specific information routing. Note that during inference, all the input is presented equally to both decoders without any dropout.

\section{Baseline HTR System}
\label{section4}
Our baseline system {\footnote{https://github.com/0x454447415244/HandwritingRecognitionSystem} follows an encoder-decoder architecture similar to the one described in \cite{chammas2018handwriting} except that we do not define an explicit sliding window. A 7-layer CNN encoder (see table \ref{table1} ) generates a sequence of observations by scanning the text-line image (normalized to 64 pixels in height) in the direction of the writing (e.g., left to right for Latin scripts). We sample the image at a rate of 1/4 (selected empirically), which means that the encoder generates one feature vector for every 4 pixels. Extracted feature vectors are fed into a 3-layer Bidirectional LSTM (BLSTM) decoder with 256 single-cell, peephole-enabled, units per layer. We chose to keep the network simple with a relatively small number of parameters. We thus combine the forward and backward outputs at the end of the BLSTM decoder rather than at each BLSTM layer. We also chose not to add additional fully-connected layers. The network is trained in an end-to-end fashion with the connectionist temporal classification (CTC) loss function \cite{graves2006connectionist}. Spatial pooling (max) is employed after some convolutional layers with a stride equal to 2 (except for the last layer). Because of the CTC limitation, we only apply pooling twice in the horizontal direction to keep the number of observations (per line) greater than the number of labels.

\begin{table}[h]

	\caption{Encoder CNN. All convolutions use the leaky ReLU as activation function (with $\alpha = 0.2$) \cite{he2015delving}, followed by a batch normalization layer \cite{ioffe2015batch}.}
	
	\label{table1}

	\centering

	\begin{tabular}{c}
		
		\toprule
		Output \\
		
		\midrule		
		4 $\times$ 1 Max pooling \\	
		\midrule
		3 $\times$ 3 Convolution, 512 features \\

		\midrule		
		2 $\times$ 1 Max pooling \\	
		\midrule
		3 $\times$ 3 Convolution, 512 features \\
		\midrule
		3 $\times$ 3 Convolution, 512 features \\
		
		\midrule
		2 $\times$ 1 Max pooling \\
		\midrule
		3 $\times$ 3 Convolution, 256 features \\
		\midrule
		3 $\times$ 3 Convolution, 256 features \\
		
		\midrule
		2 $\times$ 2 Max pooling \\
		\midrule
		3 $\times$ 3 Convolution, 128 features \\
		
		\midrule
		2 $\times$ 2 Max pooling \\
		\midrule
		3 $\times$ 3 Convolution, 64 features \\

		\midrule		
		Input \\
		
		\bottomrule

	\end{tabular}
	
\end{table}

\section{Datasets}
\label{section5}
We performed our experiments on the Rodrigo dataset \cite{serrano2010rodrigo}, a corpus obtained from the digitization of the “Historia de España del arçobispo Don Rodrigo” book, written in ancient Spanish in 1545. The dataset consists of 9000 lines for training, 1000 lines for validation and 5100 lines for testing. Other experiments were performed on the READ16 dataset \cite{sanchez2016icfhr2016}, a subset of documents from the Ratsprotokolle collection, composed of minutes of the council meetings held from 1470 to 1805. The dataset, written in Early Modern German, consists of 8360 lines for training and 1040 lines for validation. It is worth noting that raw grayscale text-line images are fed directly into the encoder network without any preprocessing.

\section{Experiments}
\label{section6}

We fine-tune our baseline system that has already converged to a sub-optimal solution by retraining it on the same data with TD applied to even-numbered mini-batches. In table \ref{table2}, we show the performance of the system on the two datasets. We notice that TD improves the performance when applied at the input layer as well as at the CNN encoder output. A combination of both approaches improves the performance even more with 15\% and 16\% reduction in relative raw label error rate on READ16 and Rodrigo datasets, respectively. The TD with Complementary Input Representation (CIR) seems to improve the performance when implemented at the decoder level as well as when used in combination with TD applied to the input image. The difference in performance between TD CIR and the baseline TD method can be interpreted as due to the RNN decoders co-adaptation with regard to the proposed information routing.

\begin{table}[h]
	
	\caption{Raw label error rate on the validation data of Rodrigo and READ16 datasets.}
	
	\label{table2}
	
	\centering
	
	\begin{tabular}{ccc}
		
		\toprule
		
		System        				& Rodrigo & READ16 						\\
		
		\midrule
		CRNN 	  	 					& 3.02	& 10.79 					\\
		CRNN + TD (Image)				& 2.75	& 10.00						\\
		CRNN + TD (Encoder)   			& 2.75  & 10.06						\\
		CRNN + TD (CIR Encoders)   		& 2.73  &  9.76						\\
		CRNN + TD (Image \& Encoder)   	& \textbf{2.50}  & \textbf{9.13}	\\
		CRNN + TD (Image \& CIR Encoders)   & {2.59}  & {9.48}				\\
		\bottomrule
		
	\end{tabular}
	
\end{table}

To assess the effect of TD, we compute the average magnitude for each of the 256 features of the RNN decoder output over all samples in the test set. We can notice that the overall features magnitudes are higher for the baseline system than for the one with TD (figure \ref{figure4}). The features in the system with TD tend to be more uniform by having lower standard deviation. We notice also that dead features (ones with low magnitude) are more active in the system with TD (figure \ref{figure5}) as it seems to make use of most features.

\begin{figure}[h]
	\centering
	\includegraphics[width=1.0\linewidth]{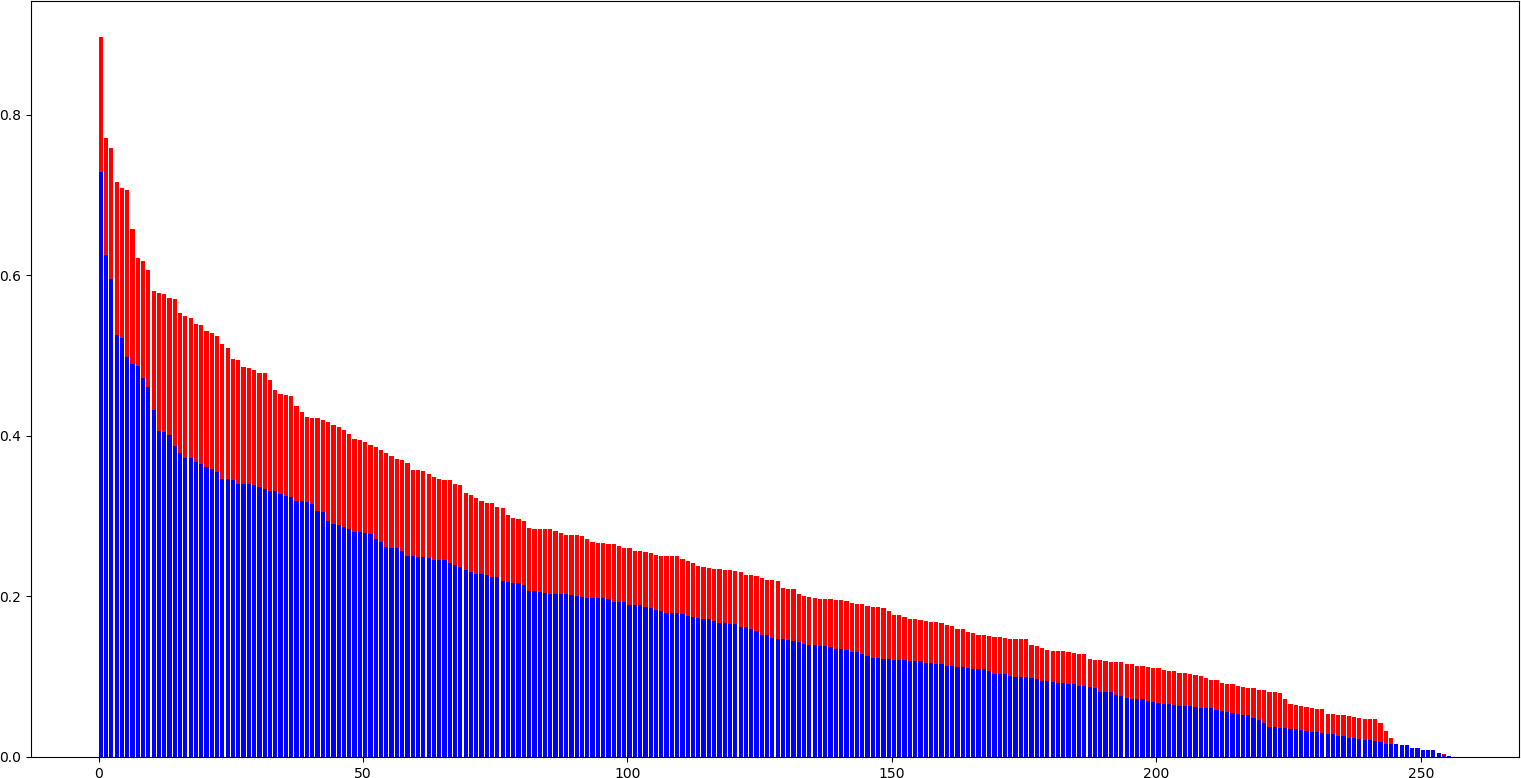}
	\caption{BLSTM decoder forward network features averaged over all samples in the READ16 validation set and sorted by their magnitude. Red bars are for the baseline system while the blue bars belong to the one with Temporal Dropout.}
	\label{figure4}
\end{figure}

\begin{figure}[h]
	\centering
	\includegraphics[width=1.0\linewidth]{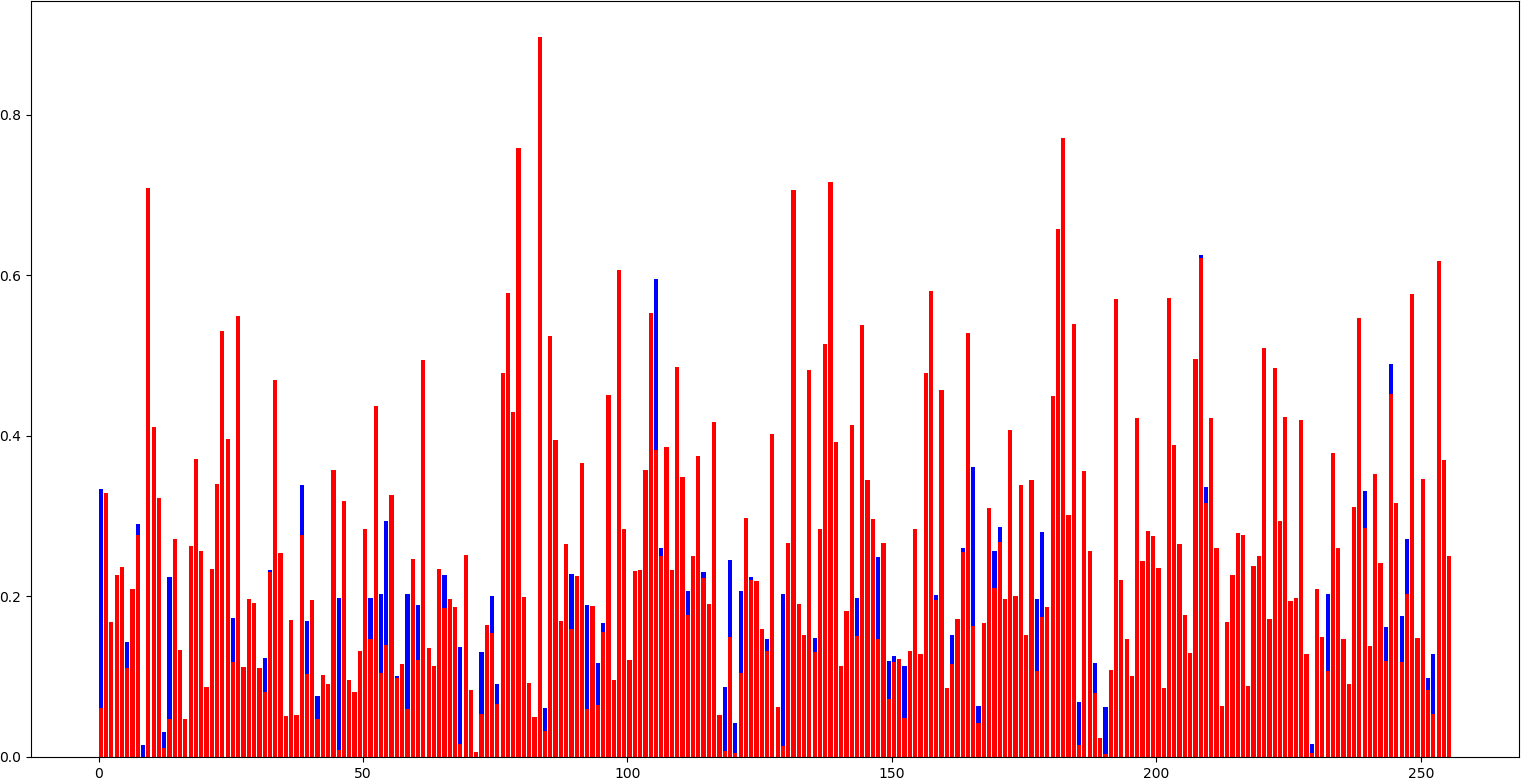}
	\caption{Unsorted version of features in figure \ref{figure4}.}
	\label{figure5}
\end{figure}

We assess the importance of features with respect to their magnitude by clipping values above a certain threshold $\gamma$. For example, with $\gamma$ equals to 0.6, all features with magnitude greater than 0.6 are considered as 0.6. In figure \ref{figure6}, we show the effect of features clipping for the two systems. We can see that the system with TD is more robust toward features clipping with less than 0.5\% loss of accuracy for a threshold of 0.3. This can be explained as the system with TD makes use of more features with low magnitude (see figures \ref{figure4} and \ref{figure5}).

\begin{figure}[h]
	\centering
	\includegraphics[width=1.0\linewidth]{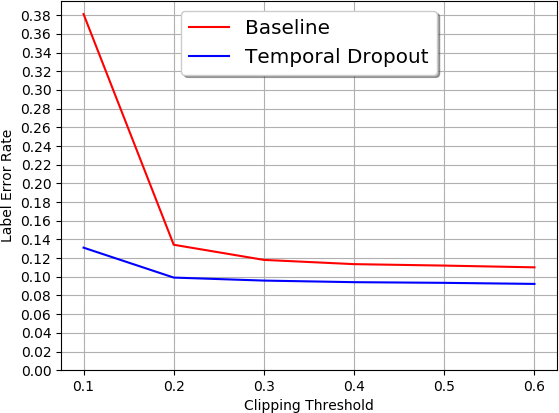}
	\caption{System performance (lower is better) on READ16 dataset with respect to different features clipping threshold values.}
	\label{figure6}
\end{figure}

To integrate a lexicon and a language model, we use Weighted Finite State Transducers (WFST). The WFST decoder is based on the CTC-specific implementation proposed by \cite{miao2015eesen} for speech recognition. A “token” WFST was designed to handle all possible label sequences at the frame level, so as to allow for the occurrence of the blank label along with the repetition of non-blank labels. It can map a sequence of frame-level CTC labels to a single character. A search graph is built with three WFSTs ($T$, $L$ and $G$) compiled independently and combined as follows:

\begin{equation}
S = T \: \circ \: min(det(L \: \circ \: G))
\end{equation}

$T$, $L$ and $G$ are the token, lexicon and grammar WFSTs respectively, whereas ,  and  denote composition, determination and minimization respectively. The determination and minimization operations are needed to compress the search space, yielding a faster decoding.

\begin{table}[h]
	
	\caption{Word and character error rates on the test data of Rodrigo dataset.}
	
	\label{table3}
	
	\centering
	
	\begin{tabular}{cccc}
		
		\toprule
		
		System     & \# parameters	& WER/CER \\
		
		\midrule
		CRNN 			&10.5 M & 16.63 / 4.70	 	\\
		CRNN + TD 		&10.5 M & 14.37 / 3.74 	\\
		Deep CRNN \cite{granell2018transcription} &18.5 M & 14.00 / 3.00  \\	
		Deep CRNN 		&18.5 M & 12.93 / 2.62		\\
		Deep CRNN + TD	&18.5 M & \textbf{12.31 / 2.43}		\\	
		\bottomrule
		
	\end{tabular}
	
\end{table}

We report the results of the different experiments in Table \ref{table3} where an 8-gram character language model is used. The system prefixed with “Deep” denote the use of 13 convolutional layers in the encoder (instead of 7), as described in \cite{chammas2018handwriting}. Besides the size of the network, one difference between the CRNN system implemented in this paper and the one in \cite{granell2018transcription} is that we opted out of using an explicit sliding window on the text-line image. The encoder now performs a 2D convolution on the image as a whole, where it generate one feature vector for each 4 pixel columns. Another difference is that we now use the leaky ReLU activation function in the encoder (instead of ReLU).

\section{Conclusion}
\label{section7}
In this paper, we proposed a novel method to improve the performance of a recurrent convolutional neural network for handwritten text recognition. Our method fine-tunes a pretrained model by applying temporal dropout to the input image as well to internal network representation. Our results show that stochastically dropping sequence elements improves modeling accuracy. Specifically, we found out that models with temporal dropout make the most out of their output features and are more robust toward feature clipping. We have also investigated the use of a complementary RNN decoder to process dropped out information. While this method improves the recognition accuracy compared to the baseline system, it suffers from decoders co-adaptation due to the constant dropout rate. Future work will investigate the use of a stochastic or learned dropout rate to mitigate this behavior.

\section{Acknowledgment}

\label{section8}

We gratefully acknowledge the support of NVIDIA Corporation with the donation of the Titan Xp GPU used for this research.

\bibliographystyle{IEEEtran}
\bibliography{ref}

\begin{thebibliography}{10}
\providecommand{\url}[1]{#1}
\csname url@samestyle\endcsname
\providecommand{\newblock}{\relax}
\providecommand{\bibinfo}[2]{#2}
\providecommand{\BIBentrySTDinterwordspacing}{\spaceskip=0pt\relax}
\providecommand{\BIBentryALTinterwordstretchfactor}{4}
\providecommand{\BIBentryALTinterwordspacing}{\spaceskip=\fontdimen2\font plus
\BIBentryALTinterwordstretchfactor\fontdimen3\font minus
  \fontdimen4\font\relax}
\providecommand{\BIBforeignlanguage}[2]{{%
\expandafter\ifx\csname l@#1\endcsname\relax
\typeout{** WARNING: IEEEtran.bst: No hyphenation pattern has been}%
\typeout{** loaded for the language `#1'. Using the pattern for}%
\typeout{** the default language instead.}%
\else
\language=\csname l@#1\endcsname
\fi
#2}}
\providecommand{\BIBdecl}{\relax}
\BIBdecl

\bibitem{hinton2012improving}
G.~E. Hinton, N.~Srivastava, A.~Krizhevsky, I.~Sutskever, and R.~R.
  Salakhutdinov, ``Improving neural networks by preventing co-adaptation of
  feature detectors,'' \emph{arXiv preprint arXiv:1207.0580}, 2012.

\bibitem{wan2013regularization}
L.~Wan, M.~Zeiler, S.~Zhang, Y.~Le~Cun, and R.~Fergus, ``Regularization of
  neural networks using dropconnect,'' in \emph{International conference on
  machine learning}, 2013, pp. 1058--1066.

\bibitem{devries2017improved}
T.~DeVries and G.~W. Taylor, ``Improved regularization of convolutional neural
  networks with cutout,'' \emph{arXiv preprint arXiv:1708.04552}, 2017.

\bibitem{tompson2015efficient}
J.~Tompson, R.~Goroshin, A.~Jain, Y.~LeCun, and C.~Bregler, ``Efficient object
  localization using convolutional networks,'' in \emph{Proceedings of the IEEE
  Conference on Computer Vision and Pattern Recognition}, 2015, pp. 648--656.

\bibitem{culibrk2014temporal}
D.~Culibrk and N.~Sebe, ``Temporal dropout of changes approach to convolutional
  learning of spatio-temporal features,'' in \emph{Proceedings of the 22nd ACM
  international conference on Multimedia}.\hskip 1em plus 0.5em minus
  0.4em\relax ACM, 2014, pp. 1201--1204.

\bibitem{pham2014dropout}
V.~Pham, T.~Bluche, C.~Kermorvant, and J.~Louradour, ``Dropout improves
  recurrent neural networks for handwriting recognition,'' in \emph{2014 14th
  International Conference on Frontiers in Handwriting Recognition}.\hskip 1em
  plus 0.5em minus 0.4em\relax IEEE, 2014, pp. 285--290.

\bibitem{gal2016theoretically}
Y.~Gal and Z.~Ghahramani, ``A theoretically grounded application of dropout in
  recurrent neural networks,'' in \emph{Advances in neural information
  processing systems}, 2016, pp. 1019--1027.

\bibitem{moon2015rnndrop}
T.~Moon, H.~Choi, H.~Lee, and I.~Song, ``Rnndrop: A novel dropout for rnns in
  asr,'' in \emph{2015 IEEE Workshop on Automatic Speech Recognition and
  Understanding (ASRU)}.\hskip 1em plus 0.5em minus 0.4em\relax IEEE, 2015, pp.
  65--70.

\bibitem{semeniuta2016recurrent}
S.~Semeniuta, A.~Severyn, and E.~Barth, ``Recurrent dropout without memory
  loss,'' \emph{arXiv preprint arXiv:1603.05118}, 2016.

\bibitem{an1996effects}
G.~An, ``The effects of adding noise during backpropagation training on a
  generalization performance,'' \emph{Neural computation}, vol.~8, no.~3, pp.
  643--674, 1996.

\bibitem{neelakantan2015adding}
A.~Neelakantan, L.~Vilnis, Q.~V. Le, I.~Sutskever, L.~Kaiser, K.~Kurach, and
  J.~Martens, ``Adding gradient noise improves learning for very deep
  networks,'' \emph{arXiv preprint arXiv:1511.06807}, 2015.

\bibitem{li2016whiteout}
Y.~Li and F.~Liu, ``Whiteout: Gaussian adaptive noise regularization in deep
  neural networks,'' \emph{arXiv preprint arXiv:1612.01490}, 2016.

\bibitem{kang2017shakeout}
G.~Kang, J.~Li, and D.~Tao, ``Shakeout: A new approach to regularized deep
  neural network training,'' \emph{IEEE transactions on pattern analysis and
  machine intelligence}, vol.~40, no.~5, pp. 1245--1258, 2017.

\bibitem{larsson2016fractalnet}
G.~Larsson, M.~Maire, and G.~Shakhnarovich, ``Fractalnet: Ultra-deep neural
  networks without residuals,'' \emph{arXiv preprint arXiv:1605.07648}, 2016.

\bibitem{huang2016deep}
G.~Huang, Y.~Sun, Z.~Liu, D.~Sedra, and K.~Q. Weinberger, ``Deep networks with
  stochastic depth,'' in \emph{European conference on computer vision}.\hskip
  1em plus 0.5em minus 0.4em\relax Springer, 2016, pp. 646--661.

\bibitem{gastaldi2017shake}
X.~Gastaldi, ``Shake-shake regularization,'' \emph{arXiv preprint
  arXiv:1705.07485}, 2017.

\bibitem{chammas2018handwriting}
E.~Chammas, C.~Mokbel, and L.~Likforman-Sulem, ``Handwriting recognition of
  historical documents with few labeled data,'' in \emph{2018 13th IAPR
  International Workshop on Document Analysis Systems (DAS)}.\hskip 1em plus
  0.5em minus 0.4em\relax IEEE, 2018, pp. 43--48.

\bibitem{graves2006connectionist}
A.~Graves, S.~Fern{\'a}ndez, F.~Gomez, and J.~Schmidhuber, ``Connectionist
  temporal classification: labelling unsegmented sequence data with recurrent
  neural networks,'' in \emph{Proceedings of the 23rd international conference
  on Machine learning}.\hskip 1em plus 0.5em minus 0.4em\relax ACM, 2006, pp.
  369--376.

\bibitem{he2015delving}
K.~He, X.~Zhang, S.~Ren, and J.~Sun, ``Delving deep into rectifiers: Surpassing
  human-level performance on imagenet classification,'' in \emph{Proceedings of
  the IEEE international conference on computer vision}, 2015, pp. 1026--1034.

\bibitem{ioffe2015batch}
S.~Ioffe and C.~Szegedy, ``Batch normalization: Accelerating deep network
  training by reducing internal covariate shift,'' \emph{arXiv preprint
  arXiv:1502.03167}, 2015.

\bibitem{serrano2010rodrigo}
N.~Serrano, F.~Castro, and A.~Juan, ``The rodrigo database.'' in \emph{LREC},
  2010, pp. 19--21.

\bibitem{sanchez2016icfhr2016}
J.~A. Sanchez, V.~Romero, A.~H. Toselli, and E.~Vidal, ``Icfhr2016 competition
  on handwritten text recognition on the read dataset,'' in \emph{2016 15th
  International Conference on Frontiers in Handwriting Recognition
  (ICFHR)}.\hskip 1em plus 0.5em minus 0.4em\relax IEEE, 2016, pp. 630--635.

\bibitem{miao2015eesen}
Y.~Miao, M.~Gowayyed, and F.~Metze, ``Eesen: End-to-end speech recognition
  using deep rnn models and wfst-based decoding,'' in \emph{2015 IEEE Workshop
  on Automatic Speech Recognition and Understanding (ASRU)}.\hskip 1em plus
  0.5em minus 0.4em\relax IEEE, 2015, pp. 167--174.

\bibitem{granell2018transcription}
E.~Granell, E.~Chammas, L.~Likforman-Sulem, C.-D. Mart{\'\i}nez-Hinarejos,
  C.~Mokbel, and B.-I. C{\^\i}rstea, ``Transcription of spanish historical
  handwritten documents with deep neural networks,'' \emph{Journal of Imaging},
  vol.~4, no.~1, p.~15, 2018.

\end{thebibliography}

\end{document}